\documentclass[10pt,twocolumn,letterpaper]{article}

\usepackage{cvpr}
\usepackage{times}
\usepackage{epsfig}
\usepackage{graphicx}
\usepackage{amsmath}
\usepackage{amssymb}
\usepackage{sidecap}
\usepackage{subfigure}
\usepackage{bm}
\usepackage{xcolor}

\usepackage[pagebackref=true,breaklinks=true,letterpaper=true,colorlinks,bookmarks=false]{hyperref}

\cvprfinalcopy 

\ifcvprfinal\fi
\begin{document}

\title{Learning to Generate Images With Perceptual Similarity Metrics}

\author{
\begin{tabular}{ccc}
    \begin{tabular}{@{}c@{}}
        Jake Snell \\ University of Toronto
    \end{tabular} &
    \begin{tabular}{@{}c@{}}
        Karl Ridgeway \\ University of Colorado, Boulder
    \end{tabular} &
    \begin{tabular}{@{}c@{}}
        Renjie Liao \\ University Toronto
    \end{tabular} \\  \\
    \begin{tabular}{@{}c@{}}
        Brett D. Roads \\ University of Colorado, Boulder
    \end{tabular} &
    \begin{tabular}{@{}c@{}}
        Michael C. Mozer\\ University of Colorado, Boulder
    \end{tabular} &
    \begin{tabular}{@{}c@{}}
        Richard S. Zemel \\ University Toronto
    \end{tabular}
\end{tabular}
}

\maketitle

\begin{abstract} 
Deep networks are increasingly being applied to problems involving image
synthesis, e.g., generating images from textual descriptions and reconstructing
an input image from a compact representation.  Supervised training of
image-synthesis networks typically uses a pixel-wise loss (PL) to indicate the
mismatch between a generated image and its corresponding target image. We
propose instead to use a loss function that is better calibrated to human
perceptual judgments of image quality: the multiscale structural-similarity
score (MS-SSIM) \cite{Wang2003_multiscalessim}. Because MS-SSIM is
differentiable, it is easily incorporated into gradient-descent learning. We
compare the consequences of using MS-SSIM versus PL loss on training
deterministic and stochastic autoencoders. For three different architectures,
we collected human judgments of the quality of image reconstructions.
Observers reliably prefer images synthesized by MS-SSIM-optimized models over
those synthesized by PL-optimized models, for two distinct PL measures
($L_1$ and $L_2$ distances). We also explore the effect of training
objective on image encoding and analyze conditions under which
perceptually-optimized representations yield better performance on image
classification. Finally, we demonstrate the superiority of perceptually-optimized
networks for super-resolution imaging. Just as computer vision has advanced through the use of
convolutional architectures that mimic the structure of the mammalian visual
system, we argue that significant additional advances can be made in modeling
images through the use of training objectives that are well aligned to
characteristics of human perception. 
\end{abstract}

\section{Introduction}

There has been a recent explosion of interest in developing methods for image representation learning, focused in particular on training neural networks to synthesize images.
The reason for this surge is threefold.  First, 
the problem of image generation spans a wide range of difficulty, from synthetic
images to handwritten digits to naturally cluttered and 
high-dimensional scenes, the latter of which provides a fertile development and testing
ground for generative models.
Second, learning good generative models of images involves learning
new representations. Such representations are believed to be useful for a variety of machine 
learning tasks,
such as classification or clustering, and can also support transfer between tasks.
They are also applicable to other vision problems, including
analysis by synthesis, learning of 3D representations, and future prediction in  video.   
Third, image generation is fun and captures popular imagination, as efforts such as Google's
Inceptionism machine demonstrate.

\begin{figure}[tb]
\begin{center}
\includegraphics[scale=0.55]{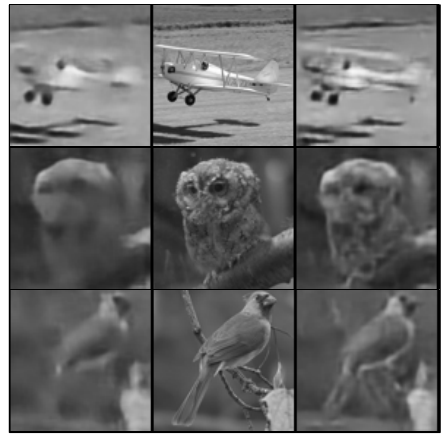}
\end{center}
\caption{Three examples showing reconstructions of an original image (center) by a standard reconstruction approach (left) and our technique (right). The compression factor is high to highlight the differences.}
\label{fig:example}
\end{figure}

While unsupervised image representation learning has become a popular task,
there is surprisingly little work on studying loss functions that are
appropriate for image generation.
A basic method for learning generative image models is the {\em autoencoder} architecture. Autoencoders are made up of two functions, an encoder and a decoder. The encoder compresses an image into a feature vector, typically of low dimension, and the decoder takes that vector as input and reconstructs the original image as output. The standard loss function is the 
squared Euclidean ($L_2$) distance between the original and reconstructed images, also referred to as the \emph{mean squared error} or \emph{MSE}. A city-block ($L_1$) distance is sometimes used as well, referred to as the \emph{mean absolute error} or \emph{MAE}. As we will show, both loss functions yield blurry results--synthesized images that appear to have been low-pass filtered. Probabilistic formulations of autoencoders have also been proposed that maximize the likelihood of the observed images being generated, but estimating likelihoods in high-dimensional image space is notoriously difficult \cite{Theis2016}.

In this paper, we explore loss functions that, unlike MSE, MAE, and likelihoods, are grounded in human perceptual judgments. We show that these perceptual losses lead to representations are superior to other methods, both with respect to reconstructing given images (Figure~\ref{fig:example}), and generating novel ones. This superiority is demonstrated both in quantitative studies and human judgements. Beyond achieving perceptually superior synthesized images, we also show that that our perceptually-optimized representations are better suited for image classification. Finally, we demonstrate that perceptual losses yield a convincing win when applied to a state-of-the-art architecture for single image super-resolution.

\section{Background and Related Work}

\subsection{Neural networks for image synthesis}

The standard neural network for image synthesis is the autoencoder, of which there are two primary types.
In the classic {\em deterministic} autoencoder, the input is mapped directly through hidden layers to output a reconstruction of the original image. The autoencoder is trained to reproduce an image that is similar to the input, where similarity is evaluated using a pixel-wise loss between the image and its reconstruction. In a  {\em probabilistic} autoencoder, the encoder is used to approximate a posterior distribution and the decoder
is used to stochastically reconstruct the data from latent variables; the model output is viewed as a distribution over images, and the model is trained to maximize the likelihood of the original image under this distribution. 
The chief advantage of probabilistic autoencoders is that they permit stochastic generation of novel images. The key issue with probabilistic autoencoders concerns the intractability of inference in the latent variables, e.g., Helmholtz Machines \cite{dayan1995helmholtz}. Variational autoencoders (VAEs)  \cite{kingma2013auto} utilize simple variational distributions to address this issue.

A second approach to building generative models for image synthesis uses variants of
Boltzmann Machines \cite{Smolensky1986,Hinton1986} and Deep Belief Networks \cite{Hinton2006}.
While these models are very powerful, each iteration of
training requires a computationally costly step of MCMC to approximate
derivatives of an intractable partition function (normalization constant),
making it difficult to scale them to large datasets.

A third approach to learning generative image models, which we refer to as the {\em direct-generation} approach,
involves training a generator that maps random samples drawn from a uniform 
distribution through a deep neural network that outputs images,
and attempts through training to make the set of
images generated by the model indistinguishable from real images.
Generative Adversarial Networks (GANs) \cite{Goodfellow2014} is a paradigm that
involves training a discriminator that attempts to distinguish real from generated images,
along with a generator that attempts to trick the discriminator.
Recently, this approach has been scaled by training
conditional GANs at each level of a Laplacian pyramid of images \cite{Denton2015}. 
With some additional clever training ideas, these adversarial networks have produced 
very impressive generative results, e.g., \cite{Radford2015}. Drawbacks of the GAN 
include the need to train a second network, a deep and complicated adversary, and the fact 
that the training of the two networks are inter-dependent and lack a single common objective.
An alternative approach, moment-matching networks \cite{Li2015}, directly
trains the generator to make the statistics of these two distributions match. 

Because the goal of image generation is to synthesize images that humans would judge as high quality and natural, current approaches seem inadequate by failing to incorporate measures of human perception. With direct-generation approaches, human judgments could in principle be incorporated by replacing the GAN with human discrimination of real from generated images. However, in practice, the required amount of human effort would make such a scheme impractical. In this paper, we describe an alternative approach using the autoencoder architecture; this approach incorporates image assessments consistent with human perceptual judgments without requiring human data collection.

We focus on autoencoders over direct-generation approaches for a second reason: autoencoders {\em interpret} images in addition to generating images. That is, an input image can be mapped to a compact representation that encodes the underlying properties of the world responsible for the observed image features.  This joint training of the encoder and decoder facilitates task transfer: the encoder can be used as the initial image mapping that can be utilized for many different applications. Although
adversarial training can be combined with autoencoding, here we explore autoencoding in isolation, to study the effects of optimizing with perceptually-based metrics.

Because autoencoders reconstruct training images, training the network requires evaluating the
quality of the reconstruction with respect to the original.
This evaluation is
based on a pixel-to-pixel comparison of the images---a so-called {\em full-reference metric}. Deterministic autoencoders typically use \emph{mean-squared error} (\textit{MSE}), the average square of the pixel intensity differences, or \emph{mean-absolute} error (\textit{MAE}), the average of the absolute difference in pixel intensity. Probabilistic autoencoders typically use a likelihood measure that is a monotonically decreasing function of pixelwise differences. In many instances, these three standard measures---MSE, MAE, and likelihood---fail to capture human judgments of quality. For example, a distorted image created by decreasing the contrast can yield the same standard measure as one created by increasing the contrast, but the two distortions can yield quite different human judgments of visual quality; and distorting an image with salt-and-pepper impulse noise obtains a small perturbation by standard measures but is judged by people as having low visual quality relative to the original image.

\subsection{Perception-Based Error Metrics}

As digitization of photos and videos became commonplace in the 1990s, the need for digital
compression also became apparent. Lossy compression schemes distorted image data, and it was
important to quantify the drop in quality resulting from compression in order to optimize
the compression scheme. Because compressed digital
artifacts are eventually used by humans, researchers attempted to develop full-reference
image quality metrics that take into account features to which the human visual system is sensitive
and that ignore features to which it is insensitive.  Some are built on
complex models of the human visual system, such as the Sarnoff JND model \cite{lubin1998human},
the visual differences predictor \cite{daly1992visible_differences_predictor},
the moving picture quality metric \cite{van1996perceptual}, the perceptual
distortion metric \cite{winkler1998perceptual}, and the metric of \cite{frese1997methodology}.

Others take more of an engineering approach, and are based on the extraction
and analysis of specific features of an image to which human perception is sensitive.
The most popular of these metrics is the structural similarity metric (SSIM) \cite{wang2004ssim}, which 
aims to match the luminance, contrast, and structure information in an image.
Alternative engineering-based metrics are the visual information fidelity metric \cite{sheikh2006image},
which is an information theory-based measure, and the visual signal-to-noise 
ratio \cite{chandler2007vsnr}.

Finally, there are transform-based methods, which compare the images after some
transformation has been applied. Some of these methods include DCT/wavelets, discrete
orthonormal transforms, and singular value decomposition. 

\subsection{Structural Similarity}

In this paper, we train neural nets with the structural-similarity metric (SSIM) \cite{wang2004ssim} and its multiscale extension (MS-SSIM) \cite{Wang2003_multiscalessim}.
We chose the SSIM family of metrics because it is well accepted and frequently utilized in the literature.
Further, its pixelwise gradient has a simple analytical form and is inexpensive to compute.
In this work, we focus on the original grayscale SSIM and MS-SSIM, although there are interesting
variations and improvements such as colorized SSIM \cite{kolaman2012quaternion,Hassan2012_colorssim}.

The single-scale SSIM metric \cite{wang2004ssim} compares corresponding pixels and their neighborhoods in two images, denoted $x$ and $y$, with three comparison functions---luminance ($I$), contrast ($C$), and 
structure ($S$):
\begin{equation*} \label{eq:ssim_components}
\begin{split}
I(x,y) \!=\! \frac{2\mu_x\mu_y + C_1}{\mu_x^2+\mu_y^2+C_1}  ~&~~~
C(x,y) \!=\! \frac{2\sigma_x\sigma_y + C_2}{\sigma_x^2+\sigma_y^2+C_2}  \\
S(x,y) &= \frac{\sigma_{xy} + C_3}{\sigma_x\sigma_y+C_3}
\end{split}
\end{equation*}
The variables $\mu_x$, $\mu_y$, $\sigma_x$, and $\sigma_y$ denote
mean pixel intensity and the standard deviations of pixel intensity in a local image patch centered at either $x$ or
$y$. Following \cite{wang2004ssim}, we chose a square neighborhood of 5 pixels on either side of $x$ or $y$, resulting
in $11\times11$ patches. 
The variable $\sigma_{xy}$ denotes the sample correlation
coefficient between corresponding pixels in the patches centered at $x$ and $y$. The constants $C_1$, $C_2$, and $C_3$ 
are small values added for numerical stability. The three comparison functions
are combined to form the SSIM score:
\vspace{-.035in} 
\begin{equation*}\label{eq:ssim_1}
\text{SSIM}(x,y) = I(x,y)^\alpha  C(x,y)^\beta  S(x,y)^\gamma
\vspace{-.035in}
\end{equation*}
This single-scale measure assumes a fixed image sampling density and viewing distance, and may only
be appropriate for certain range of image scales. This issue is addressed in \cite{Wang2003_multiscalessim}
with a variant of SSIM that operates at multiple scales simultaneously.
The input images $x$ and $y$ are iteratively downsampled by a factor of 2 with a low-pass filter, with scale $j$ denoting the original images downsampled by a factor of $2^{j-1}$. The contrast $C(x,y)$ and
structure $S(x,y)$ components are applied at all scales. The luminance component is applied only at the
coarsest scale, denoted $M$. Additionally, a weighting is allowed for the contrast and structure components at each scale, leading to the definition:
\vspace{-.035in}
\begin{equation*}\label{eq:mssim}
\text{MS-SSIM}(x,y) = I_M(x,y)^{\alpha_M}  \prod_{j=1}^{M}{ C_j(x,y)^{\beta_j}  S_j(X,y)^{\gamma_j}  }
\vspace{-.035in}
\end{equation*}
In our work, we weight each component and each scale equally ($\alpha = \beta_{1..M} = \gamma_{1..M} =1$), a common simplification of MS-SSIM. Following \cite{Wang2003_multiscalessim}, we use $M=5$ downsampling steps. 

Our objective is to minimize the loss related to the sum of structural-similarity scores across
all image pixels,
\vspace{-.05in}
\begin{equation*}\label{eq:objfn}
\mathcal{L}(X,Y) = -\sum_{i} \text{MS-SSIM}(X_i,Y_i) ,
\vspace{-.05in}
\end{equation*}
where $X$ and $Y$ are the original and reconstructed images, and $i$ is an index over image pixels. This equation
has a simple analytical derivative \cite{wang2008ssim_derivative}
and therefore it is trivial to perform gradient descent in the MS-SSIM-related loss.

We now turn to two sets of simulation experiments that compare autoencoders trained with a pixelwise loss (MSE and MAE) to those trained with a perceptually optimized loss (SSIM or MS-SSIM). The first set of experiments is based on deterministic autoencoders, and the second is based on a probabilistic autoencoder, the VAE \cite{kingma2013auto}.

\section{Deterministic Autoencoders}
We demonstrate the benefits of training deterministic autoencoders to optimize SSIM or MS-SSIM 
across images of various sizes, for various bottlenecks in the autoencoder, and for various network architectures.
We begin with a study using small, highly compressed images and 
a fully connected architecture. We then present results on larger images with a convolutional autoencoder architecture. 

\subsection{Architectures and Data Sets}
\label{sec:detarch}

In the first simulation, we trained networks on small ($32 \times 32$) images using a fully-connected architecture with a six-layer encoder mapping the 1024-dimensional input to a bottleneck layer of 256 units. The decoder component of the architecture mirrors the encoder. We trained two networks that are identical except for their loss function---one to optimize MSE, and one to optimize SSIM. Because the images are so small, the single-scale SSIM is appropriate; downsampling the images any further blurs the content to the point where humans have trouble distinguishing objects in the image. We train the fully-connected autoencoders using a subset of approximately two million images of the 80 million Tiny-Images data set \cite{tinyimages}, consisting of the first 30 images for every English proper noun. The $32 \times 32$ images in this dataset consist of RGB color channels. We mapped the three color channels to a single grayscale channel using the ITU-R 601-2 luma transform. All testing and evaluation of our models used the CIFAR-10 data set, which consists of 60,000 color images, each drawn from one of ten categories. We chose a diverse data set for training in order ensure that the autoencoders were learning general statistical characteristics of images, and not peculiarities of the CIFAR-10 data set. The CIFAR-10 color images were converted to a single grayscale channel, as was done for the training data set. Additional details regarding the architecture and training procedure can be found in section~\ref{sec:fcae_details}.

Next we trained networks on larger images ($96 \times 96$ pixels) with a {\em convolutional} autoencoder architecture \cite{masci2011conv_autoencoders}: convolutional layers encode the input and deconvolutional layers decode the feature representation in the bottleneck layer. The convolutional network architecture consists of 3 convolutional layers, each with a filter size of 5  and a stride of 2. The deconvolutional layers again mirror the convolutional layers. For these larger images, which may have structure at multiple spatial scales, we used the MS-SSIM rather than SSIM as our perceptual similarity metric. We compared MS-SSIM to two pixelwise measures: MSE and MAE. Because MSE focuses on outliers and we have no reason to believe that the human eye has a similar focus, we felt it important to include MAE. If we observe MS-SSIM outperforming both MSE and MAE, we will have stronger evidence for the conclusion that perceptually-optimized measures outperform pixelwise losses in general. For training and testing, we use the STL-10 dataset \cite{stl10}, which consists of larger RGB color images, of the same classes as CIFAR-10. The images were converted to grayscale using the method we used for CIFAR-10. For our experiments, we train our models on the 100,000 images in STL-10 referred to as the ``unlabeled'' set, and of the remaining data, we formed a {\em validation} set of 10,400 images and a {\em test} set of 2,800 images.

\subsection{Judgments of Reconstruction Quality}

Do human observers prefer reconstructions produced by perceptually-optimized networks or by the pixelwise-loss optimized networks? We collected judgments of perceptual quality on Amazon Mechanical Turk.

\subsubsection{Fully-Connected Autoencoders}
Participants were shown image triplets with the original (reference) image in the center and the SSIM- and MSE-optimized reconstructions on either side with the locations counterbalanced. Participants were instructed to select which of the two reconstructions
they preferred. 

\begin{figure}[bt]
\begin{center}
\includegraphics[width=1.6in]{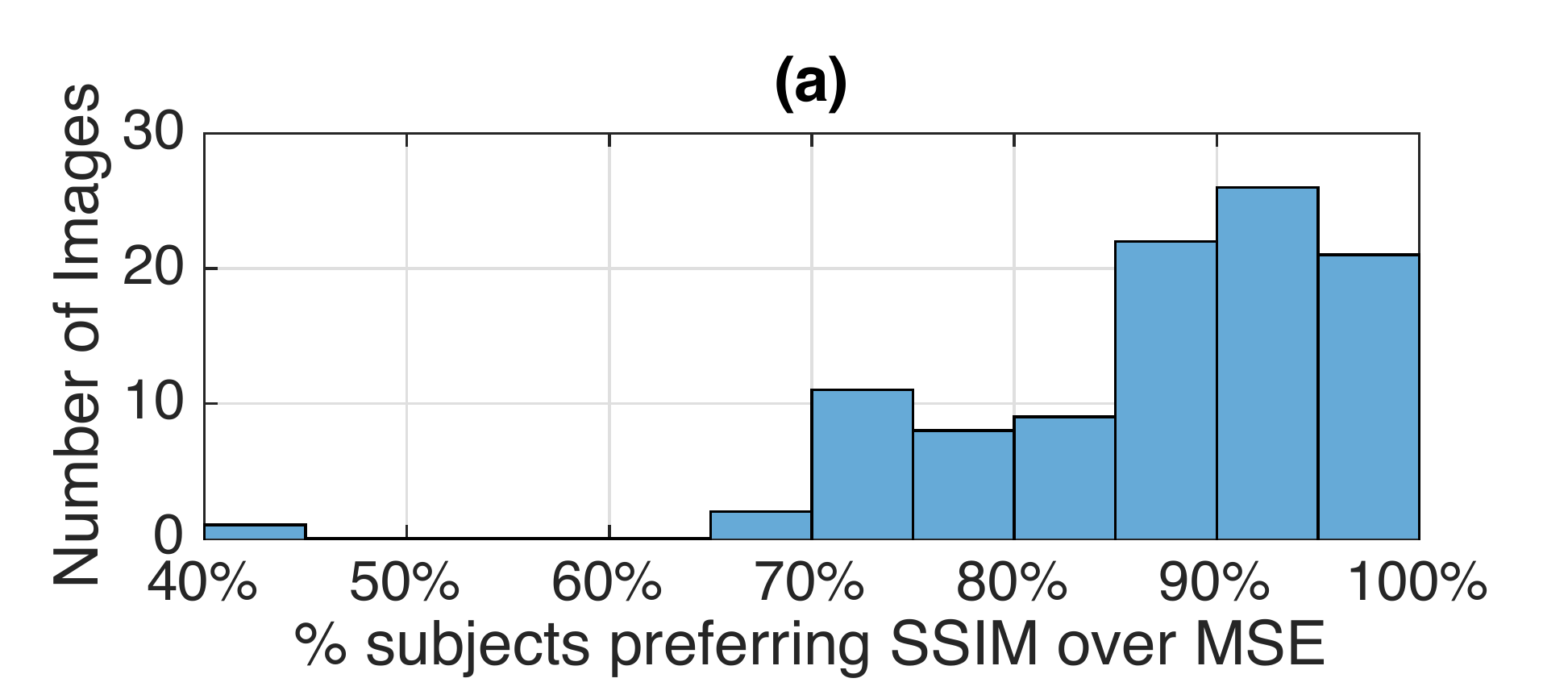}
\includegraphics[width=1.6in]{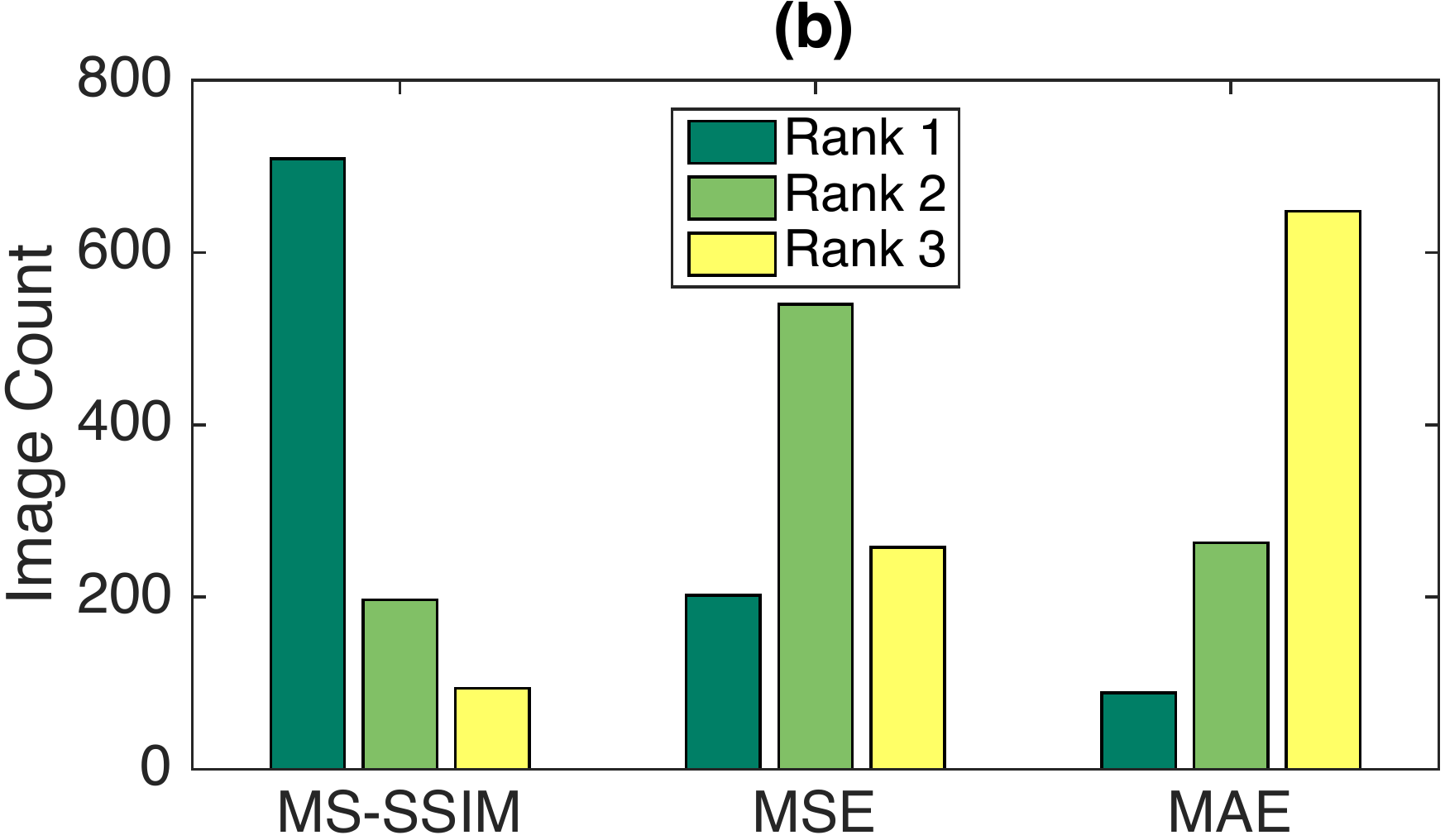}
\end{center}
\caption{Human judgments of reconstructed images. (a) Fully connected network: Proportion of participants preferring SSIM to MSE for each of 100 image triplets. (b) Deterministic conv. network: Distribution of image quality ranking for MS-SSIM, MSE, and MAE for 1000 images from the STL-10 hold-out set.}
\label{fig:human_rankings}
\end{figure}

In a first study, twenty participants provided preference judgments on the same set of 100
randomly selected images from the CIFAR-10 data set. For each image triple, we recorded the proportion of participants who choose the SSIM reconstruction of the image over the MSE reconstruction. Figure~\ref{fig:human_rankings}a
shows the distribution of inter-participant preference for SSIM reconstructions across all 100 images.
If participants were choosing randomly, we would expect to see roughly 50\% preference
for most images. However,  a plurality of images have over 90\% inter-participant 
agreement on SSIM, and almost no images have MSE reconstructions that are preferred over SSIM reconstructions
by a majority of participants.

\begin{figure}[bt]
\centering
(a) \includegraphics[scale=0.45]{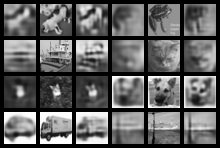}\hspace{.1in}
(b) \includegraphics[scale=0.45]{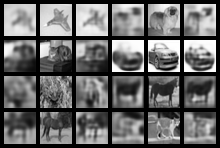}
\caption{Image triples consisting of---from left to right---the MSE reconstruction, the original image, and the SSIM reconstruction. 
Image triples are ordered, from top to bottom
and left to right, by the percentage of participants preferring
SSIM. (a) Eight images for which participants strongly preferred SSIM over MSE.
(b) Eight images for which the smallest proportion of participants preferred SSIM. }
\label{fig:ssim_reconstructions}
\end{figure}

Figure~\ref{fig:ssim_reconstructions}a shows the eight image triplets for which the largest
proportion of participants preferred the SSIM reconstruction. The original image
is shown in the center of the triplet and the MSE- and SSIM-optimized reconstructions appear on the
left and right, respectively. (In the actual experiment, the two reconstructions were flipped on
half of the trials.) The SSIM reconstructions all show important object details
that are lost in the MSE reconstructions and were unanimously preferred by participants.
Figure~\ref{fig:ssim_reconstructions}b shows the eight image triples for which the
smallest proportion of participants preferred the SSIM reconstruction. In the first seven of
these images, still a majority (70\%) of participants preferred the SSIM reconstruction to
the MSE reconstruction; only in the image in the lower right corner did a majority prefer the
MSE reconstruction (60\%). The SSIM-optimized reconstructions still seem to show as
much detail as the MSE-optimized reconstructions, and the inconsistency in the ratings may
indicate that the two reconstructions are of about equal quality.

In a second study on Mechanical Turk, twenty new participants each provided preference judgments on a randomly drawn
set of 100 images and their reconstructions. The images were different for each participant; consequently,
a total of 2000 images were judged. Participants preferred the SSIM- over MSE-optimized reconstructions by nearly a
7:1 ratio: the SSIM reconstruction was chosen for 86.25\% of the images. 
Individual participants chose SSIM reconstruction between 63\% and 99\% of trials.

\subsubsection{Convolutional Autoencoders} 
We also performed a third Mechanical Turk study, this time on the convolutional autoencoders, to determine whether human observers prefer images generated by the MS-SSIM-optimized networks to MSE- and
MAE-optimized networks. Images were chosen randomly from the STL-10 validation set.  Participants were presented with a sequence of screens showing the original (reference) image on the left and a set of of three reconstructions on the right. Participants were instructed to drag and drop the images vertically into the correct order, so that the best reconstruction is on top and the worst on the bottom. The initial vertical ordering of reconstructions was randomized.
We asked 20 participants to each rank 50 images, for a total of 1000 rankings.
Figure~\ref{fig:human_rankings}b shows the distribution
over rankings for each of the three training objectives. If participants chose randomly,
one would expect to see the same number of high rankings for each model.
However, MS-SSIM is ranked highest for a majority of images (709 out of 1000).

\begin{figure}[bt]
\centering
(a) \includegraphics[width=.4\linewidth]{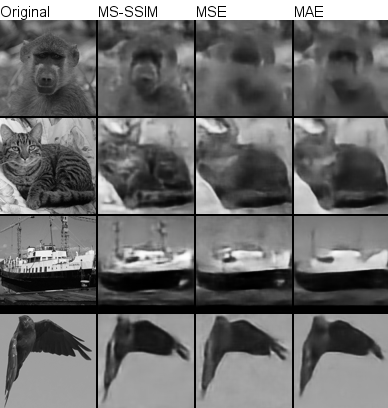} 
(b) \includegraphics[width=.4\linewidth]{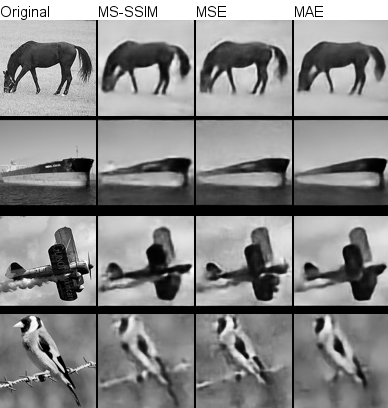} 
\caption{ (a) Four randomly selected, held-out STL-10 images and their
reconstructions for the 128-hidden-unit networks. 
For these images, the MS-SSIM reconstruction was ranked as best by humans.
(b)  Four randomly selected test images where the MS-SSIM reconstruction was ranked second or third.
}
\label{fig:128_autoencoder_rank_examples}
\end{figure}

Figure~\ref{fig:128_autoencoder_rank_examples}a shows  examples
of images whose MS-SSIM reconstruction was ranked as best by human judges.
Figure~\ref{fig:128_autoencoder_rank_examples}b shows examples
of images whose MSE or MAE reconstruction was ranked as the best.
The strong preference for MS-SSIM appears to be due to its superiority
in capturing fine detail such as the monkey and cat faces and background detail such as
the construction cranes. MS-SSIM seems to have less of an advantage on
simpler, more homogeneous, less textured images. Note that even
when MSE or MAE beats MS-SSIM, the MS-SSIM reconstructions have no obvious
defects relative to the other reconstructions.

\section{Probabilistic Autoencoders}
\label{sec:elVAE_qual}

In order to further explore the role of perceptual losses in learning models for image generation, we adapt the variational autoencoder (VAE) model of \cite{kingma2013auto} to be trained with an arbitrary differentiable image similarity metric. The VAE closely resembles a standard autoencoder, utilizing a combination of an encoding network that produces a code for an image $x$, and then a decoding network that maps the code to an image $\hat{x}$. The key difference is that the code $z$ is considered a latent variable, endowed with a prior $p(z)$. The encoder $q_{\bm{\phi}}(z|x)$, parameterized by $\bm{\phi}$, approximates the intractable posterior of $z$ given the image $x$. The decoder $p_{\bm{\theta}}(x | z)$, parameterized by $\bm{\theta}$, produces a distribution over images given $z$. The VAE minimizes a variational upper bound on the negative log-likelihood of the data:
\vspace{-.035in}
\begin{equation*}
\mathcal{L}^{VAE} = \mathbb{E}_{q_{\bm{\phi}}} [-\log p_{\bm{\theta}}(x|z) ] + D_{KL}(q_{\bm{\phi}}(z | x) || p(z))
\vspace{-.035in}
\end{equation*}
We modify this learning objective to better suit an arbitrary differential loss $\Delta(x, \hat{x})$
by replacing the probabilistic decoder with a deterministic prediction as a function of the code:
$\hat{x} \equiv f_{\bm{\theta}}(z)$. The objective then becomes a weighted sum of the expected loss of $\hat{x}$
under the encoder's distribution over $z$ and the KL regularization term:
\vspace{-.035in}
\begin{equation}
    \mathcal{L}^{EL} =  C \cdot \mathbb{E}_{q_{\bm{\phi}}} \left[\Delta(x, \hat{x})\right] + D_{KL}(q_{\bm{\phi}}(z | x) || p(z))
    \label{eq:elVAE_obj}
\vspace{-.035in}
\end{equation}
where the constant $C$ governs the trade-off between the image-specific loss and the regularizer. We call this
modification {\em Expected-Loss VAE} (EL-VAE).

\subsection{EL-VAE Training Methodology}
\label{sec:elVAE_hyper}

We trained convolutional EL-VAE networks with 128-dimensional $z$ on 96 $\times$ 96 pixel images from the unlabeled portion of STL-10 with a similar architecture to the deterministic convolutional autoencoders. One key choice when training EL-VAEs is the value of $C$ in Equation~\ref{eq:elVAE_obj}, which governs the trade-off between the KL loss and reconstruction error. As $C$ increases, the model will put greater emphasis on reconstructions. At the same time, the KL-divergence of the prior from the approximate posterior will increase, leading to poorer samples. Selecting a value of $C$ is further complicated due to the different scaling depending on the choice of the image-specific loss $\Delta$.

In order to mitigate the differences in scaling, we normalized each loss (MSE, MAE, and MS-SSIM) by dividing
by its expected value as estimated by computing the loss on 10,000 pairs randomly drawn with replacement
from the training set. To select the best value of $C$, we utilized a recent approach to model selection in
generative models \cite{bounliphone2016}. This work proposes a statistical test of relative similarity to determine
which model generates samples that are significantly closer to the reference dataset of interest. The test statistic is the difference in squared maximum mean discrepancies (MMDs) between the reference dataset and a dataset generated by each model. We trained convolutional EL-VAEs with $C \in \{1, 10, 1000, 10000\}$ for each loss on a 5,000 example subset of the STL-10 unlabeled dataset. We then utilized the test statistic of \cite{bounliphone2016} to determine for each loss the value of $C$ that produced samples with smallest squared MMD compared to the STL-10 train set. For each loss $C = 1000$ was selected by this test and thus we used this value when training EL-VAEs on the full unlabeled STL-10 dataset.

\subsection{EL-VAE Results}

\begin{SCfigure}[.4][bt]
\includegraphics[width=2.0in]{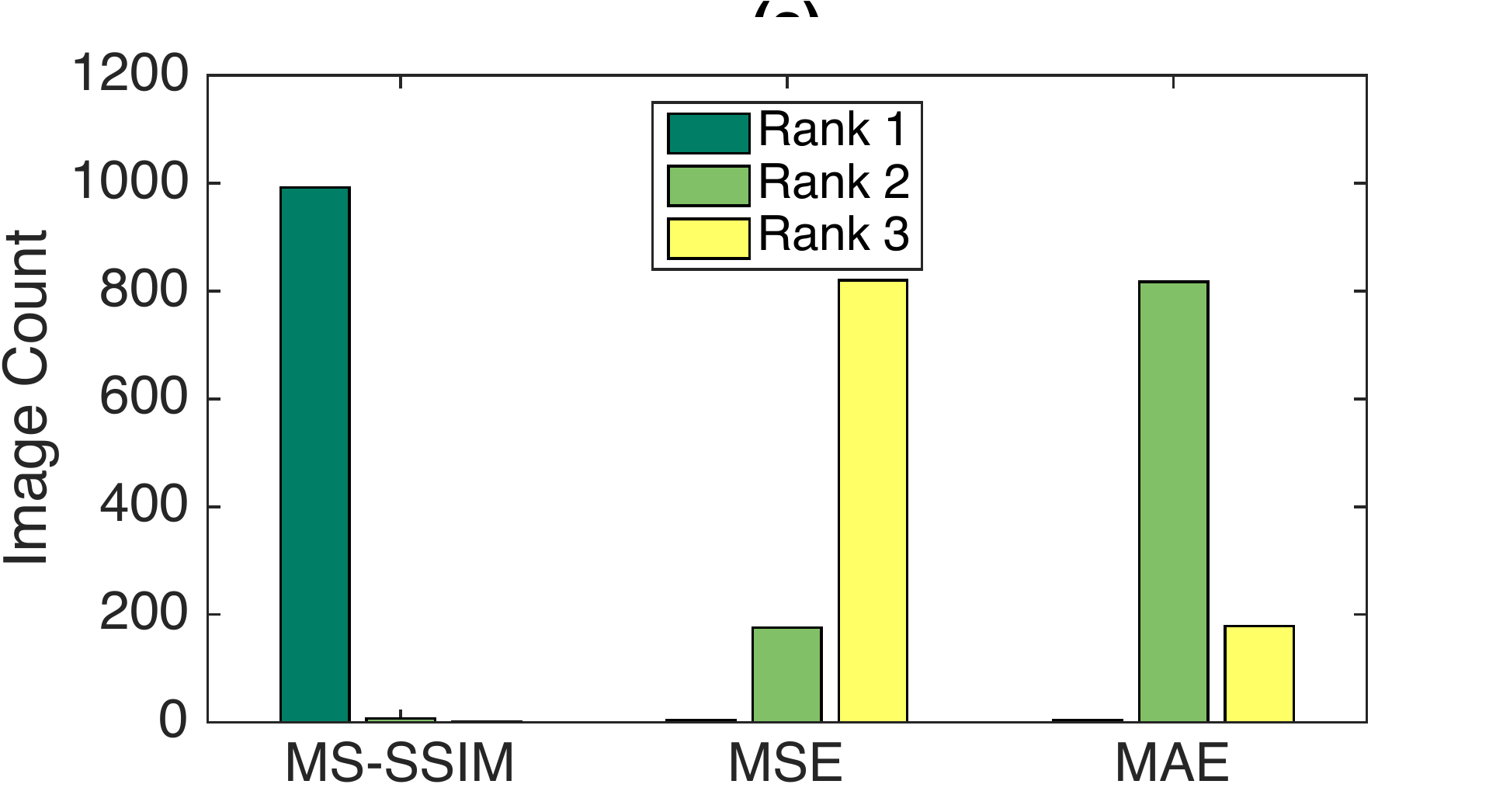}
\caption{Image quality ranking for MS-SSIM, MSE-, and MAE-optimized EL-VAEs.}
\label{fig:elhuman_rankings}
\end{SCfigure}

\begin{figure}[bt]
    \centering
    (a)
    \includegraphics[width=.4\linewidth]{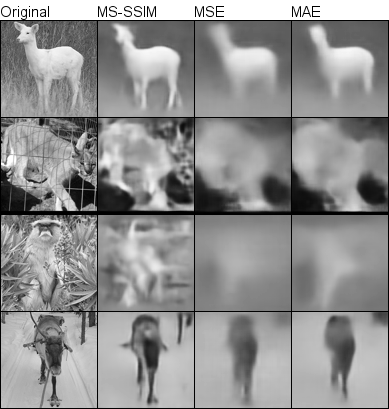} 
    (b)
    \includegraphics[width=.4\linewidth]{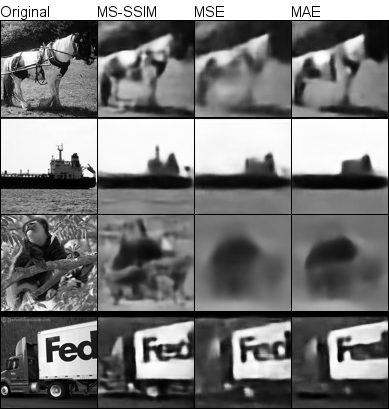} 
    \caption{ (a) Four randomly selected, held-out STL-10 images and their
    reconstructions. For these images, the MS-SSIM reconstruction was ranked as best by humans. Reconstructions are from the 128-hidden-unit VAEs. From left to right are the original image, followed by the MS-SSIM, MSE, and MAE reconstructions.
    (b) Four randomly selected test images where the MS-SSIM reconstruction was ranked second or third.    \label{fig:128_vae_recons}
}
\end{figure}

We performed a final Mechanical Turk study to determine human observer preferences for image reconstructions generated by MS-SSIM-, MSE-, and MAE-optimized EL-VAE architecture.  We generated reconstructions of 1000 randomly chosen images from the STL-10 test set by taking the latent code to be the mode of the approximate posterior for each EL-VAE network. The procedure was otherwise the same as detailed above for 
the deterministic case. MS-SSIM was ranked the highest in 992 out of 1000 cases. Figure~\ref{fig:elhuman_rankings} shows the distribution of image rankings for each loss. Test reconstructions for the EL-VAE networks are shown in Figure~\ref{fig:128_vae_recons}. Figure~\ref{fig:128_vae_recons}a shows reconstructions for which MS-SSIM was ranked as best, and Figure~\ref{fig:128_vae_recons}b shows reconstructions for which it was not ranked best. As observed in the deterministic case, MS-SSIM is better at capturing
fine details than either MSE or MAE.

\begin{figure*}[bt]
    \begin{center}
        \includegraphics[width=0.8\linewidth]{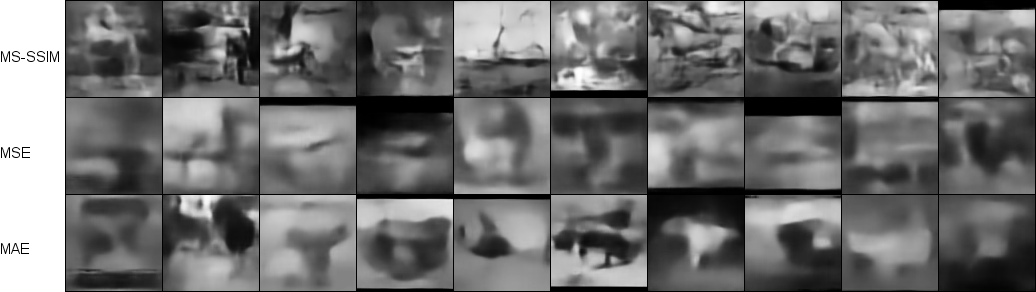}
    \end{center}
    \caption{Samples generated from EL-VAEs optimized with MS-SSIM (top row), MSE (middle row) and MAE (bottom
    row). Sample quality appears to mirror reconstruction quality: the MS-SSIM optimized EL-VAE generates fine details that the other models do not. }
    \label{fig:vae_samples}
\end{figure*}

In order to qualitatively assess the performance of each EL-VAEs as a generative model, Figure~\ref{fig:vae_samples} shows
random samples from each model. Each image was generated by drawing a code $z$ from the prior and then passing it
as input to the decoder. The samples generated by the MS-SSIM-optimized net contain a great degree of detail and
structure. 

\section{Classification with Learned Representations}

In the previous sections, we showed that using a perceptually-aligned training objective improves the quality of image synthesis, as judged by human observers, for three different neural net architectures. In this section, we investigate whether the MS-SSIM objective leads to the discovery of internal representations in the neural net that are more closely tied to the factors of variation in images. For these experiments we use the Extended Yale B Faces dataset~\cite{lee2005acquiring}. This dataset contains 2,414 grayscale images of 38 individuals and is labeled with the azimuth ($-130^\circ$ to $+130^\circ$) and elevation ($-40^\circ$ to $+90^\circ$) of the light source in relation to the face. We resized the images to $48 \times 48$ and split the data randomly into a training, validation, and test set in a 60\%-20\%-20\% ratio. We learned deterministic convolutional autoencoders using MSE, MAE, and MS-SSIM as loss functions and then used the bottleneck representations as features for SVMs trained to predict identity, azimuth, and elevation. We opted to investigate this prediction task as opposed to a more straightforward task (such as STL-10 classification accuracy) because we expect MS-SSIM to obtain superior encodings of low- and mid-level visual features such as edges and contours. Indeed, as predicted, initial studies showed only modest benefits of MS-SSIM for STL-10 classification accuracy, where coarse classification (e.g., plane versus ship) does not require fine image detail.  

The deterministic convolutional autoencoders we trained on Yale B had a similar architecture to those described in Section~\ref{sec:detarch}. Here though we used a 32-unit bottleneck layer with ReLU activations, and used batch normalization~\cite{ioffe2015batch} on all layers except the output layer of the decoder. After training each of the autoencoders to convergence on the training set, we extracted bottleneck representations for the training and validation sets. We trained a SVM with a linear kernel to predict identity and SVR with RBF kernels to predict azimuth and elevation. Hyperparameters of the SVMs were selected via three-fold cross-validation on the training plus validation set. The resulting performance on the test set (Table~\ref{tab:yalebclassification}) demonstrate that MS-SSIM yields robust representations of relevant image factors
and thereby outperforms MSE and MAE.

\begin{table}[bt]
\small
\centering
\begin{tabular}[b]{ | c | c | c | c | c | } 
\hline
Loss & Identity & Azimuth & Elevation \\ \hline
MSE & 5.60\% & 277.46 & 51.46 \\
MAE & 5.60\% & 325.19 & 50.23 \\
MS-SSIM & \textbf{3.53\%} & \textbf{234.32} & \textbf{35.60} \\
\hline
\end{tabular}
\caption{Test error for SVMs trained on bottleneck representations of deterministic convolutional autoencoders for Yale B. Classification error is the evaluation metric for identity prediction; MSE is the evaluation metric for azimuth and elevation prediction. }
\label{tab:yalebclassification}
\end{table}

\section{Image Super-Resolution}

We apply our perceptual loss to the task of super-resolution (SR) imaging. As a baseline model, we use a state-of-the-art SR method, the SRCNN \cite{dong2016image}. 
We used the SRCNN architecture determined to perform best in \cite{dong2016image}. It consists of 3 convolutional layers and 2 fully connected layers of ReLUs, with 64, 32, and 1 filters in the convolutional layers, from bottom to top, and filter sizes 9, 5, and 5. All the filters coefficients are initialized with draws from a zero-mean Gaussian with standard deviation 0.001. 

\begin{figure*}[t]
\centering
\begin{minipage}[t]{0.2050\linewidth}
\centering
\includegraphics[width=1.0\linewidth]{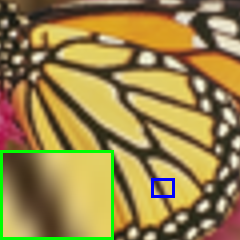}
\end{minipage}
\begin{minipage}[t]{0.2050\linewidth}
\centering
\includegraphics[width=1.0\linewidth]{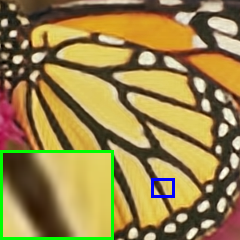}
\end{minipage}
\begin{minipage}[t]{0.2050\linewidth}
\centering
\includegraphics[width=1.0\linewidth]{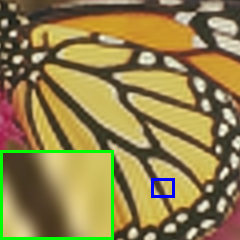}
\end{minipage}
\begin{minipage}[t]{0.2050\linewidth}
\centering
\includegraphics[width=1.0\linewidth]{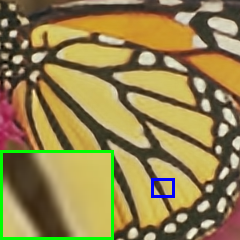}
\end{minipage} \\ 
\begin{minipage}[t]{0.2050\linewidth}
\centering
\includegraphics[width=1.0\linewidth]{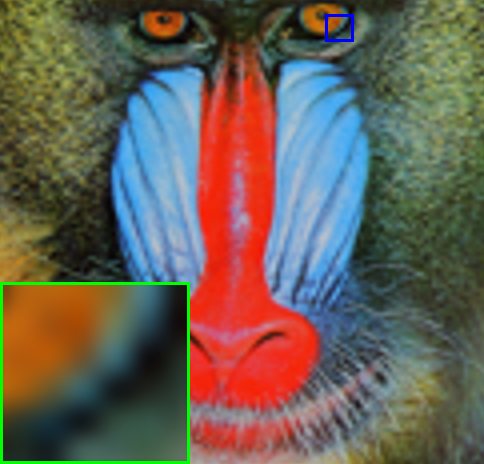}
\end{minipage}
\begin{minipage}[t]{0.2050\linewidth}
\centering
\includegraphics[width=1.0\linewidth]{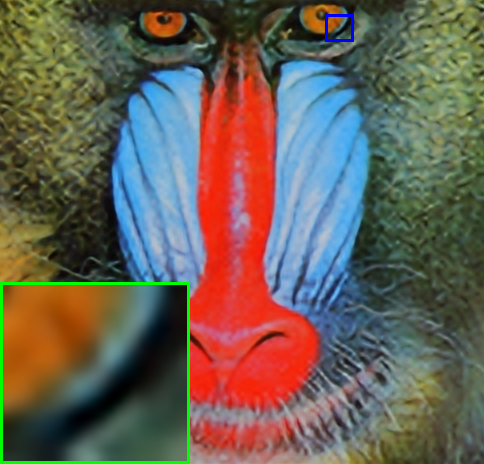}
\end{minipage}
\begin{minipage}[t]{0.2050\linewidth}
\centering
\includegraphics[width=1.0\linewidth]{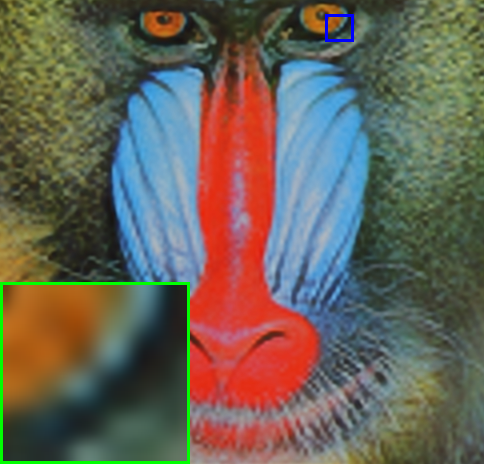}
\end{minipage}
\begin{minipage}[t]{0.2050\linewidth}
\centering
\includegraphics[width=1.0\linewidth]{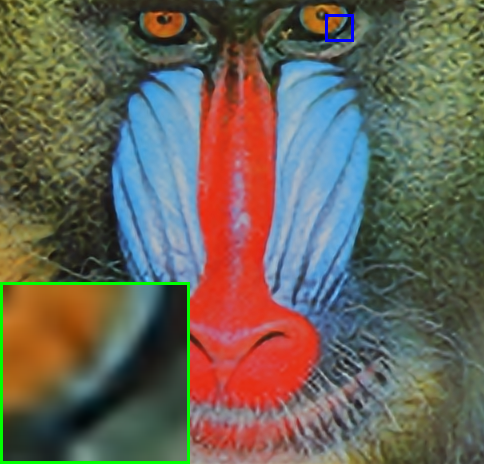}
\end{minipage} \\ 
\begin{minipage}[t]{0.2050\linewidth}
\centering
\includegraphics[width=1.0\linewidth]{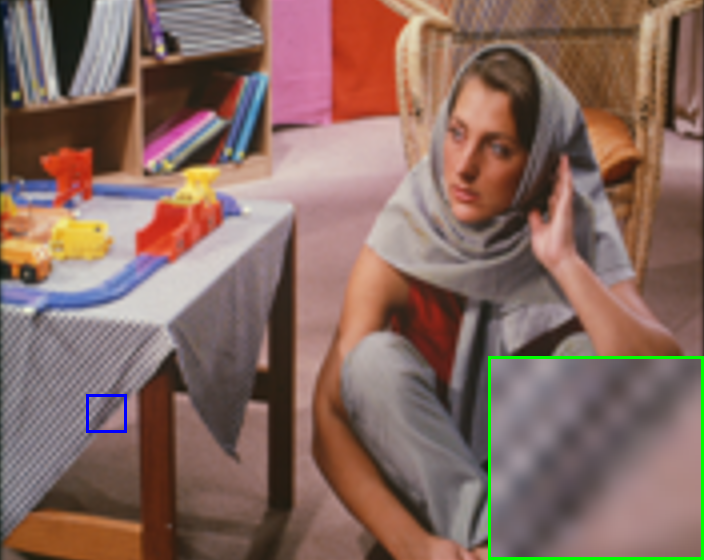}
\end{minipage}
\begin{minipage}[t]{0.2050\linewidth}
\centering
\includegraphics[width=1.0\linewidth]{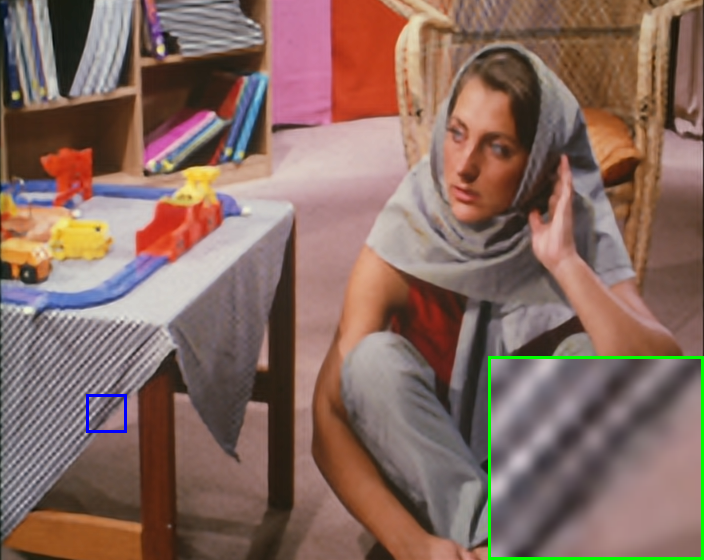}
\end{minipage}
\begin{minipage}[t]{0.2050\linewidth}
\centering
\includegraphics[width=1.0\linewidth]{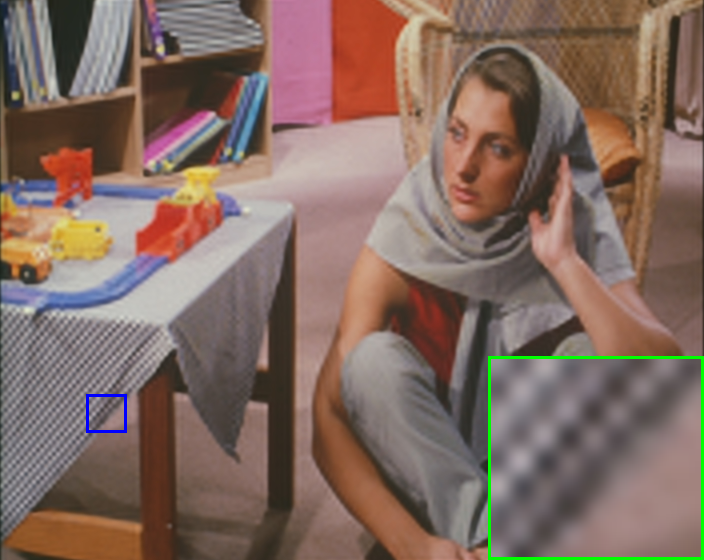}
\end{minipage}
\begin{minipage}[t]{0.2050\linewidth}
\centering
\includegraphics[width=1.0\linewidth]{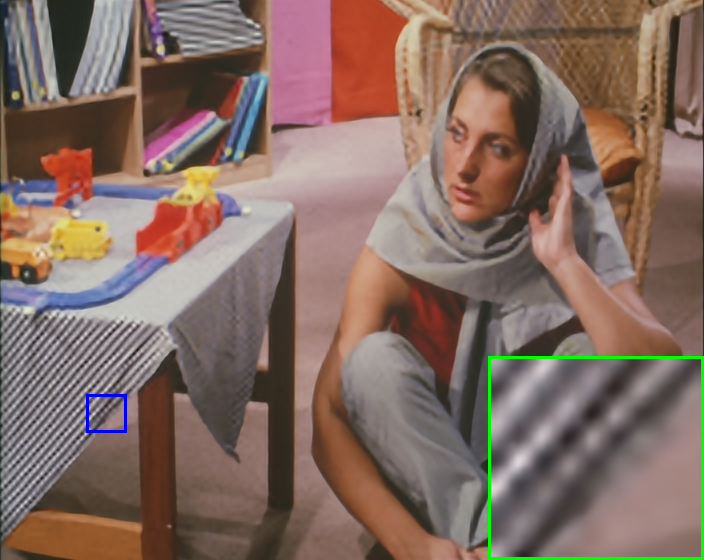}
\end{minipage} \\ 
\begin{minipage}[t]{0.2050\linewidth}
\centering
\includegraphics[width=1.0\linewidth]{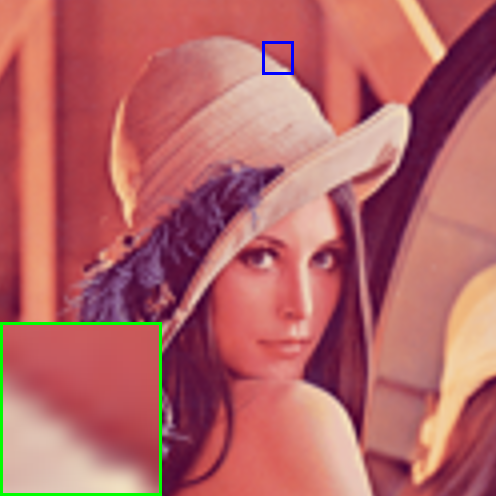}
{Bicubic}
\end{minipage}
\begin{minipage}[t]{0.2050\linewidth}
\centering
\includegraphics[width=1.0\linewidth]{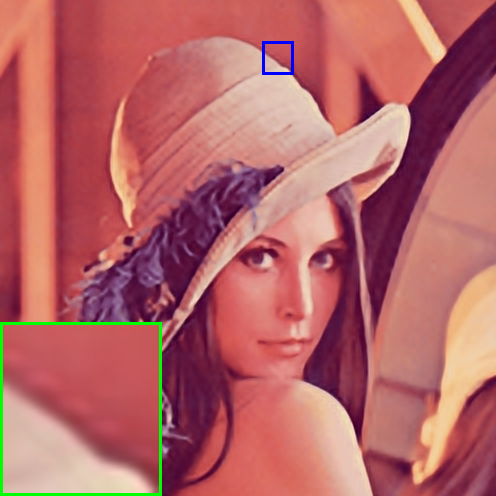}
{SRCNN + MSE}
\end{minipage}
\begin{minipage}[t]{0.2050\linewidth}
\centering
\includegraphics[width=1.0\linewidth]{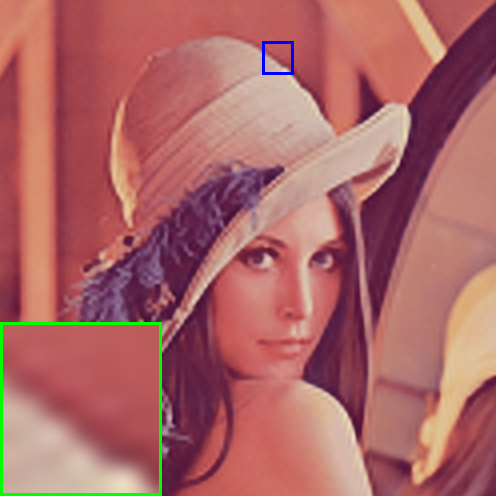}
{SRCNN + MAE}
\end{minipage}
\begin{minipage}[t]{0.2050\linewidth}
\centering
\includegraphics[width=1.0\linewidth]{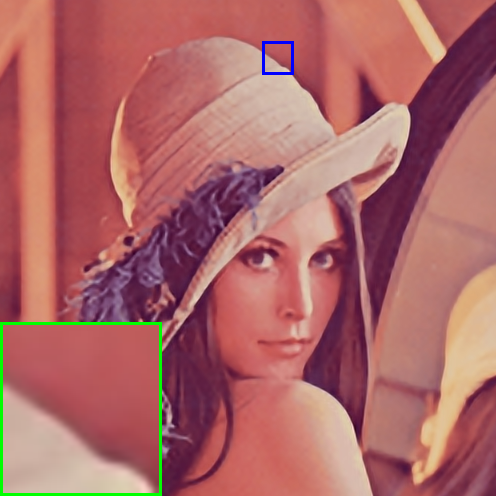}
{SRCNN + MS-SSIM}
\end{minipage} \\ 
\vspace{0.1in}
\caption{Visual comparisons on super-resolution at a magnification factor of 4.
MS-SSIM not only improves resolution but also removes artifacts, \eg, the ringing effect in the bottom row, and enhances contrast, \eg, the fabric in the third row.
\vspace{-.1in}
}
\label{fig_single_SR}
\end{figure*}

We construct a training set in a similar manner as \cite{dong2016image} by randomly cropping 5 million patches (size $33 \times 33$) from a subset of the ImageNet dataset of \cite{deng2009imagenet}. 
We compare three different loss functions for the SRCNN: MSE, MAE and MS-SSIM. Following \cite{dong2016image}, we evaluate the alternatives utilizing the standard metrics PSNR and SSIM. We tested $4\times$ SR with three standard test datasets---Set5 \cite{bevilacqua2012low}, Set14 \cite{zeyde2010single} and BSD200 \cite{martin2001database}. All measures are computed on the Y channel of YCbCr color space, averaged over the test set. 
As shown in Table \ref{tbl_sr}, MS-SSIM achieves a PSNR comparable to that of MSE, and outperforms other loss functions significantly in the SSIM measure. Fig. \ref{fig_single_SR} provides close-up visual illustrations.

\begin{table}[t]
\begin{center}
\begin{tabular}{r | c c c c }
 & Bicubic & MSE & MAE & MS-SSIM \\
\hline \hline
\textit{\footnotesize SET5}
PSNR & 28.44 & \textbf{30.52} & 29.57 & 30.35 \\
SSIM & 0.8097 & 0.8621 & 0.8350 & \textbf{0.8681} \\ \hline
\textit{\footnotesize SET14}
PSNR & 26.01 & \textbf{27.53} & 26.82 & 27.47 \\
SSIM & 0.7018 & 0.7512 & 0.7310 & \textbf{0.7610} \\ \hline
\textit{\footnotesize BSD200}
PSNR & 25.92 & \textbf{26.87} & 26.47 & 26.84 \\
SSIM & 0.6952 & 0.7378 & 0.7220 & \textbf{0.7484} \\ \hline
\end{tabular}
\caption{Super-resolution imaging results.}
\vspace{-.25in}
\label{tbl_sr}
\end{center}
\end{table}

\section{Discussion and Future Work}
We have investigated the consequences of replacing pixel-wise loss functions, MSE and MAE, with 
perceptually-grounded loss functions, SSIM and MS-SSIM, in neural networks that synthesize and transform images.
Human observers judge SSIM-optimized images to be of higher quality than PL-optimized images over a
range of neural network architectures. We also found that perceptually-optimized representations are better suited for predicting content-related image attributes. Finally, our promising results on single-image super-resolution highlight one of the key strengths of perceptual losses: they can easily be applied to current state-of-the-art architectures by simply substituting in for a pixel loss such as MSE.
Taken together, our results support the hypothesis that the MS-SSIM loss encourages networks to encode relevant low- and mid-level structure in images. Consequently, we conjecture that the MS-SSIM trained representations may even be useful for fine-grained classification tasks such as bird species identification, in which small details are important. 

\vspace{-1pt}
A recent manuscript \cite{zhao2015l2} also proposed using SSIM and MS-SSIM as a training objective for image processing neural
networks. In this manuscript, the authors evaluate alternative training objectives based not on human judgments, 
but on a range of image quality metrics. They find that MAE outperforms MSE, SSIM, and MS-SSIM on their collection of metrics, and not surprisingly, that a loss which combines both PL and SSIM measures does best---on the collection of metrics which include PL and SSIM measures. Our work goes further in demonstrating that perceptually-grounded losses attain better scores on the definitive assessment of image quality: that registered by the human visual cortex.

\vspace{-1pt}
Given our encouraging results, it seems appropriate to investigate other perceptually-grounded loss functions. SSIM is the low-hanging fruit because it is differentiable. Nonetheless, even black-box loss functions can be cached into a {\em forward model} neural net \cite{jordan1992forward} that maps image pairs into a quality measure. We can then back propagate through the forward model to transform a loss derivative expressed in perceptual quality into a loss derivative expressed in terms of individual output unit activities. This flexible framework will allow us to combine multiple  perceptually-grounded loss functions. Further, we can refine any perceptually-grounded loss functions with additional data obtained from human preference judgments, such as those we collected
in the present set of experiments.

\subsection*{Acknowledgements}
This research was supported by NSF grants SES-1461535, SBE-0542013, and SMA-1041755.

{\small
\bibliographystyle{ieee}
\bibliography{bibliography}
}
\newpage \clearpage
\appendix

\section{Deterministic Autoencoder Details}

In this section we present additional details regarding the deterministic autoencoder experiments. We also present training results demonstrating how well each network optimizes the corresponding loss function.

\subsection{Fully-Connected Autoencoders}
\label{sec:fcae_details}

Fully-connected autoencoders were trained on 32 $\times$ 32 images. The first layer expands the 1024-dimensional input to 8192 features, and each successive hidden layer reduces the dimensionality by a factor of two until we reach the bottleneck layer, which has 256 features. The decoder component of the architecture mirrors the encoder. 
In order to enforce a strong compression of the signal in our autoencoders, we force the activations of
units in the bottleneck layer to be binary ($-1$ or $+1$). Following
\cite{krizhevsky2011}, we threshold the activations in the forward pass and use the 
original continuous value for the purpose of gradient calculation during back propagation. 
We perform this quantization both during training and testing and all reported results are 
based on the quantized bottleneck-layer representations. 
All layers in the model have ReLU activations, except the bottleneck and output layers which have a tanh activation function.

The input pixels are rescaled to the range $[-1,1]$, to match the tanh activation function on all of our
output layers. We divided the CIFAR-10 images into a {\em search database} (48,000 images) and a {\em query list}
(12,000 images).  The search database was used as a validation set to determine when to stop training: training terminated when the reconstruction error---as measured by the appropriate training metric, either MSE or SSIM---stopped
improving following one complete pass through the training set.
We train using mini-batches of size 64. The SSIM and MSE metrics are scaled differently, so we performed
empirical explorations to set the learning rate appropriately for each. For MSE, 
we use a learning rate of $5\times10^{-5}$, and for SSIM $5\times10^{-2}$. All architectures were trained with 
a momentum of $0.9$ and with weight decay of $5\times10^{-5}$.

\subsection{Convolutional Autoencoders}
\label{sec:convae}

Convolutional autoencoders were trained on larger images (96 $\times$ 96 pixels). The convolutional network architecture consists of 3 convolutional layers, each with a filter size of 5 and a stride of 2. This means the input is reduced from (96,96) to (48,48), then to (24,24), and finally
to (12,12). The first layer has 128 filters, the second 256, and the final 512. All layers
have ReLU activations, except the bottleneck and output, which are tanh.
The deconvolutional layers are implemented as convolutional layers that are preceded by an upsampling step that creates a layer with 2 times the dimensions of the input layer by repeating the values of the input.
To explore the role of the capacity of the convolutional layer,
we built models with bottlenecks of both 128 and 512 units. However, we report only results for the 128-unit network
as qualitative performance of the 512-unit network was quite similar.

We train our convolutional autoencoders on the 100,000 images in STL-10 referred to as the ``unlabeled'' set, and of the remaining data, we formed a \emph{validation} set of 10,400 images and a \emph{test} set of 2,800 images. The validation set is used to determine when to stop training. Other than early stopping, no regularization was 
used during training. We used the ``Adam'' optimizer \cite{kingma2014adam}.

\subsection{Training Results}

As expected, each network performs best on its own training metric. For the fully-connected architecture, the MSE-optimized network achieves a better reconstruction MSE on $69.71\%$ of  the held-out images and the SSIM-optimized network achieves a better reconstruction SSIM score on  $97.33\%$ of images. For the convolutional architecture, we show held-out reconstruction losses for each autoencoder in Table~\ref{table:stl_recon_loss}. 

\begin{table}[h]
\small
\centering
\begin{tabular}[b]{ | c | c | c | c | c | } 
\hline
   \textbf{Loss} & \textbf{MSE} & \textbf{MAE} & \textbf{MS-SSIM} \\ \hline
MSE & \textbf{0.0233} &          0.1012 & 0.5212  \\
MAE &          0.0262 & \textbf{0.0964} & 0.5178  \\
MS-SSIM &      0.0271 &          0.1064 & \textbf{0.6018} \\
\hline
\end{tabular}
\caption{Reconstruction losses for deterministic convolutional autoencoders. The left-most column indicates the loss function used to optimize each autoencoder.}
\label{table:stl_recon_loss}
\end{table}

\section{Probabilistic Autoencoder Details}

For the EL-VAE encoders, we again used 3 convolutional layers, again with a filter size of 5 and a stride of 2. 
The number of filters matched the deterministic case: 128 in the first layer, 256 in the
second layer, 512 in the final convolutional layer, and a bottleneck of size 128.
 We used a Gaussian MLP \cite{kingma2013auto} to form the approximate posterior distribution. We chose $\mathcal{N}(0, \mathbf{I})$ as the prior. The deconvolutional layers again mirrored the convolutional layers and we used tanh units at the output.
 
We trained convolutional EL-VAEs on the unlabeled portion of STL-10 until performance on the STL-10 training
set no longer improved. No regularization was used during EL-VAE training, and the ``Adam'' optimizer 
\cite{kingma2014adam} was again used.

\section{Classification Details}

For the classification experiments, we trained deterministic convolutional autoencoders on the Extended Yale B Faces dataset~\cite{lee2005acquiring} with the same architecture described in Section~\ref{sec:convae} except here we used a 32-unit bottleneck with ReLU activations and batch normalization~\cite{ioffe2015batch} on all layers except the output layer of the decoder. Batch sizes of 32 and Adam~\cite{kingma2014adam} with a learning rate of $0.001$ were used to train the networks. For azimuth and elevation prediction, we use SVR with RBF kernels and use grid search to select $C \in \{ 1, 10, 100, 1000\}$. For identity classification, we use SVM with a linear kernel and select $C \in \{0.01, 0.1, 1, 10, 100, 1000, 10000\}$. The grid search was performed via three-fold cross-validation on the training plus validation set.

\section{Image Super-Resolution Details}

For the super-resolution experiments, all input images are converted from RGB to YCbCr color space and only Y channel is used for training and testing. To generate the training and testing data, we use bicubic method to perform downsampling. For visualizing color images, we first upsample the Cb and Cr channels using bicubic, merge the result of Y channel and then convert it back to RGB color space. The above procedure is the same as SRCNN \cite{dong2016image} which is the common practice in single image super-resolution literature. Moreover, we use stochastic gradient descent (SGD) with momentum for optimization. The learning rate is fixed to $1.0e^{-3}$ and momentum is $0.9$. We did not use any weight decay since the model is fairly simple and we did not observe any overfitting. For training the MS-SSIM loss, we use 5 scales and downsample the image with ratio 2 for each one. 

\end{document}